# Responsible Facial Recognition and Beyond

Yi Zeng[1,2,3], Enmeng Lu[1], Yinqian Sun[1], Ruochen Tian[1]

1. Institute of Automation, Chinese Academy of Sciences, China
2. Research Center for AI Ethics and Safety, Beijing Academy of Artificial Intelligence, China
3. Berggruen Research Center, Peking University, China

    Email: yi.zeng@ia.ac.cn

## Abstract

Facial recognition is changing the way we live in and interact with our society. Here we discuss the two sides of facial recognition, summarizing potential risks and current concerns. We introduce current policies and regulations in different countries. Very importantly, we point out that the risks and concerns are not only from facial recognition, but also realistically very similar to other biometric recognition technology, including but not limited to gait recognition, iris recognition, fingerprint recognition, etc. To create a responsible future, we discuss possible technological moves and efforts that should be made to keep facial recognition (and biometric recognition in general) developing for social good.

## The Two Sides of Facial Recognition

Like other technologies that change the world and the way we live, the original motivation for facial recognition is for human good. In the area of public safety, facial recognition technology has been widely used in surveillance systems, tracking criminals and identifying fugitives (Moon 2018; Lo 2018). It has also been used to fight human trafficking, detect kidnappers, and help to trace long-missing children for family reunion (Jenner 2018; Yan 2019; Cuthbertson 2018). In business and finance, facial recognition is becoming a more and more popular choice in payment and courier services, and helps to maximize security and minimize fraud (A. Lee 2017; Xia 2019; Roux 2019a). In transportation, facial recognition has been deployed in airports and train stations to save travelers time from checking in, help travelers to pay for their fares, and identify unlicensed drivers and jaywalkers (Liao 2018; 2019; Yi 2017; Tao 2018). In the medical field, facial recognition has been used in patients identification and monitoring, sentiment analysis, and genetic disorder diagnose (Martinez-Martin 2019; Roux 2019b; Vincent 2019a). In education, facial recognition helps to improve campus security, combat school bullying, as well as attendance tracking, etc. (Levy 2010; Chronicle 2018; Durkin 2019).

Although facial recognition has or will benefit our society in many ways, controversy and concerns are rising. The topic of privacy, security, accuracy, bias and freedom are with frequent discussions:

*Privacy*: When it comes to various concerns about facial recognition, privacy and data security is constantly mentioned. In February, 2019, SenseNets, a facial recognition and security software company in Shenzhen, was identified by security experts as having a serious data leak from an unprotected database, including over 2.5 million of records of citizens with personal information (Gevers 2019). In August, 2019, the personal information of over 1 million people, including biometric information such as facial recognition information and fingerprints, was found on a publicly accessible database used by the likes of UK metropolitan police, defense contractors and banks (Taylor 2019). Such data breaches can put victims at a considerable disadvantage, especially when considering the biometric information is almost permanent, and the consequences of the leak are severe and lasting.

*Security*: Although typically considered as a means of security identification, facial recognition should not be considered safe enough. Research has shown that GAN-generated Deepfakes videos are challenging for facial recognition systems, and such a challenge will be even greater when considering the further development of face-swapping technology (Korshunov and Marcel 2018). In another research, the best public Face ID system ArcFace is attacked by adding printed paper stickers on a hat and the Face ID model got confused.

*Accuracy*: In fact, when considering applying facial recognition systems in real-world scenarios, facial recognition systems might not seem to be that reliable. A report shows that the facial recognition system from South Wales Police of UK has misidentified thousands of trials, making 2,297 false positive of a total 2,470 matches, with an error rate of about 92%. Critics worry that such poor performance might result in mistaken arrests as well as a drag on the work of the police (Fingas 2018). Another evaluation from the University of Essex has shown that the Metropolitan Police's facial recognition technology only made 8 correct in its 42 matches, with an error rate of 81%, and such deployment was likely to be found "unlawful" if challenged in court (Hall 2019; Manthorpe and Martin 2019; Booth 2019).

*Bias*: Besides the robustness and reliability problems, the possible bias from or amplified by the deployment of facial recognition systems bring in extra ethical issues. In the "Gender Shades" project from MIT Media Lab and Microsoft Research, facial analysis algorithms from IBM, Microsoft, and Megvii (Face++) have been evaluated, and it shows that darker-skinned females are the most vulnerable group to gender misclassification, with error rates up to 34.4% higher than those of lighter-skinned males (Buolamwini and Gebru 2018). In a report from American Civil Liberties Union (ACLU), a test was made on Amazon's facial recognition tool "Rekognition" by comparing photos of 535 members of US Congress with a face database of 25,000 arrest photos. The results include 28 false matches, in which 39% are people of color, even though they only make up 20% of the input (Snow 2018). Another test in 2019 showed similar results (Ehrenkranz 2019). Although Amazon responded that the confidence threshold used in those tests should be set to higher values to match their recommendations for law enforcement scenarios (Wood 2018), it still raised concerns about possible harms from racial biases and whether such facial recognition tools are accurate and reliable enough for deployment (Fussell 2018; Singer 2018).

*Freedom:* Facial recognition powered surveillance systems, if improperly deployed or secured, will not only fail to effectively safeguard public safety, but also may infringe on people' freedom/privacy and provide a source for abuse. Thus, the balance between public safety and personal privacy/freedom is quite essential. In a recent survey conducted by Ipsos, about two-thirds of adults across 26 countries support "a limited and restricted government use of AI and facial recognition to maintain order", while only less than one third of citizens support "a government use as much as needed, even at the cost of sacrificing their privacy", and only less than one quarter support "a total ban" (Boyon 2019). Although the survey shows subtle differences between countries, genders and education backgrounds, it reveals a major finding that there does exist a consensus of prudence and openness about facial recognition technology.

As a case study, the recent controversy about school usage of facial recognition in China has attracted a lot of attention and discussion. The China Pharmaceutical University is reported to bring in facial recognition software for student attendance tracking and behaviour monitoring in class (Smolaks 2019). Meanwhile, a photo taken from an event went viral online, in which a demo product from Megvii, a major facial recognition company illustrated how their product could monitor and analyze students' behaviour in class, including how often they raise their hands, or lean over the table (Runhua 2019). Similar attempts are also seen in other countries (e.g. applications from the SensorStar Labs in New York, US (Alcorn 2013)). The two incidents in China quickly raised ethical concerns about current facial recognition applications in class. Some have criticized such application as invading students' privacy and freedom, some have questioned the school's ability to keep the students' facial data secure and isolate from possible abuse, some worry that this may hamper the development of students' personality and trigger a backlash from students towards AI technology, and some see such application as failure examples of empowering education with AI. The Ministry of Education of China took actions very quickly, and is thus planning to curb and regulate the use of facial recognition in schools (Siqi 2019).

Facial recognition in classrooms still needs extensive discussions on potential impacts. The motivation of the application tries to recognize whether the students are focused, tired, bored, enthusiastic, etc. which is very problematic since human emotions are very complex, and they may not be really understood by analyzing points and relations on faces. The application of this technology also changed the original way of more natural interactions between teachers and students. The results of emotion recognition in the classroom are like mirror images for students, and by using this technology, the focus of the teachers could be no longer on understanding students by interacting with them, instead, understandings are supported by statistics of these mirror images. On the other hand, there would be high risks that students may act before cameras, making fake emotional faces, and fundamentally hate technologies of such type. In addition, very possibly, similar techniques could be used to analyze the emotions, focus, and behaviors of teachers in the classrooms. And potential negative effects also apply to them. Based on the idea from Confucius, in Analects (c. 500 BCE) that "What you do not want others to do to you, do not do unto others" (similar descriptions are in Golden Rule or ethic of reciprocity), facial and emotion recognition in classrooms should not be supported. This kind of trying has a high risk to be considered as a misuse of facial recognition technology and is not responsible especially for future generations. It is also possible that it could be used in factories, offices, etc. to track and analyze behaviors of many people. Such fears are not unfounded when considering that Amazon has already been using AI systems to track warehouse workers' productivity and automatically generate paperwork to fire those that failed to meet expectations (Bort 2019), and it is reported that some sanitation workers in Nanjing, China received messages like "please continue working" from their location-tracking bracelets when they decided to stay in one place (Hollister 2019). The trying as such will very likely to have negative impacts for creating human-AI symbiotic society.

**Policies and Regulations**

In the United States, legislators in San Francisco have voted unanimously to ban the use of facial recognition technology across local agencies, including transport authority and law enforcement, making them the first of US to do so (D. L. Lee Dave 2019; Paul 2019). They argue that the ban would protect them from possible inaccuracy and bias, and maintain their privacy and liberty. A few months later, the city of Somerville and Oakland also passed their own ban on city use of facial recognition (Wu 2019; Fisher 2019). Before that, on March 14, 2019, a bipartisan bill, called the Commercial Facial Recognition Privacy Act, was introduced by senators to offer legislative oversight on the commercial application of facial recognition. The bill would prohibit commercial users from collecting and re-sharing facial data for identifying or tracking consumers without their consent (Hatmaker 2019).

In the European Union, the General Data Protection Regulation (GDPR), enforced on May 25, 2018, has already offer EU a strict regulation for the protection of personal information. The European Commission is also planning to impose strict limits on facial recognition usage to give EU citizens explicit rights over the use of their facial data (Khan 2019). It worth noting that the Swedish Data Protection Authority (DPA) has recently issued its first GDPR fine against a trial project in a school of northern Sweden, in which 22 students were captured using facial recognition software to keep track of their attendance in class (Euronews 2019; Hanselaer 2019). The Swedish DPA accused the school of processing personal data more than necessary and without legal basis, data protection impact assessment, and prior consultation (Edvardsen 2019).

In China, privacy and personal data security have become the focus of supervision since such concerns became real problems. Taking effect on June 1, 2017, the China Internet Security Law banned online service providers from collecting and selling users' personal information without user consent (Creemers, Triolo, and Webster 2018). On May 1, 2018, the national standard on personal information protection, the Personal information security specification, took effect and laid out guidelines for how personal information should be collected, used, and shared (Shi et al. 2019).

Regulations of Facial recognition are directly related to data and governance of AI in general. In China, as the leading organization that pushes forward the Chinese Next Generation Artificial Intelligence initiative, the Ministry of Science and Technology established the "Governance Principles for the New Generation Artificial Intelligence"(NGPNGAI 2019), including principles of Fairness and Justice, Respect for Privacy, etc. Relevant Efforts from Office of the Cyberspace Administration of China have also been established, such as "Data security management measures" and "Regulations for the Protection of Children's Personal Information on the Network" (CAC 2019a; 2019b), in which the importance of personal data protection, including the protection of personal biometric information, is highlighted (Xie 2019; Luo, Yu, and Shepherd 2019).

The responsible development of facial recognition does not only rely on governments regulation, but also on the active involvement from industry. Microsoft actively called for governments regulation to regulate facial recognition as well as corporate responsibility, by starting to adopt six facial recognition principles (Brad Smith 2018b; 2018a). It is reported that they denied police facial recognition project and deleted their public face recognition dataset (Vincent 2019b; Murgia 2019). Google has chosen to not offer general-purpose facial recognition APIs to avoid possible abuse and harmful outcome (Google 2018). Axon (formerly Taser International), a major police body camera manufacturer, has established their own AI Ethics Board and published the first report, concluded that the deployment of facial recognition technology in police body cameras should be stopped until such technology performs better accuracy and "equally well across races, ethnicities, genders, and other identity groups" (Axon 2019). Megvii, a major facial recognition company in China, has also set up a committee to oversee AI ethics-related issues (Sarah Dai 2019).

Alipay, the world's largest mobile payment platform, has recently published a "biometrics user privacy and security protection initiative", calling for a "minimum and sufficient" principle when collecting user biometric data, and advocating that biometric technology should be "normative and controllable" (Shen 2019).

**Similar Risks in Other Biometric Recognition**

Facial recognition, as one type of biometric recognition technology, is not all that we should care about. In fact, similar risks exist in almost every type of biometric recognition, including but not limited to gait recognition, iris recognition, fingerprint recognition, and voice recognition.

When it comes to gait recognition, we are also faced with similar risks and ethical issues. According to a study from the University of California and Beihang University (Zhang, Wang, and Bhanu 2010), ethnics can be detected using human's gait. They can distinguish people from East Asia and South America with about 80% for accuracy only based on their gaits. Moreover, even in 2005, gender classification has achieved over 95% for accuracy using gait, as suggested in the work of Dankook University and Southampton University(Yoo, Hwang, and Nixon 2005). In summary, gait recognition can also extract features like gender and ethnics just as facial recognition does, which could bring a series of problems related to algorithmic bias.

Iris recognition has similar risks. Evaluations have shown that gender, eye color, race will have a different impact on the accuracy in iris recognition. Recognition on the UND iris database shows that the accuracy on male is 96.67%, while only 86% on female (Tapia, Perez, and Bowyer 2014). An evaluation on different algorithms also shows that some of them perform better on male data, while some of them perform better on female data (Quinn et al. 2018). For eye color, 13 algorithms perform better on dark eyes (brown and black), while the rest of 27 algorithms perform better on light eyes (blue, green and grey). Concerning race, the accuracy for white people is the best, while for Asian people is the worst (Quinn et al. 2018). In (Howard and Etter 2013), similar results were presented. The false rejection rates for different races are African American > Asian > Hispanic > Caucasian. While the false rejection rates for eye colors are Black > Brown > Blue > Green > Hazel > Blue-Green (Howard and Etter 2013).

For fingerprint recognition, the accuracy on gender recognition could reach at least the accuracy of 97% (Gornale, Patil, and Veersheety 2016), which means that one can extract the gender information, and this information could have similar risks to be used with gender discrimination. In addition, the recognition accuracy for male and female are 91.69% and 84.69%, and future efforts are needed to make a balance (Wadhwa, Kaur, and Singh 2013).

**For a Responsible Future**

The future of facial recognition and related biometric recognition in society does not rely only on how people see the risks of these technologies from a societal perspective, and provide principles, norms, standards, policies, and laws, etc. to regulate them, but also heavily rely on what kind of technological move that could be made to provide a responsible future. At least the following efforts should be conducted from the technological point of view:

1. *Constantly Improving the Accuracy and Robustness of recognition models.* Many current facial recognition models are with high accuracy on restricted datasets, while performing very differently in real-world and complex scenarios, and when facing adversarial attacks. The negative impacts are especially unacceptable in safety and security related issues. With human in the loop to make final decisions, continuous efforts are needed to increase the accuracy and robustness of the models and services that are put to practical use.

2. *Upgrading Data Privacy and Security Infrastructures.* More secure data infrastructures with data grading, access control, data auditing, and privacy protection need to be developed, updated, and deployed to minimize the risks, especially for biometric data. Data firewalls need to be built against external attacks and data leaks.

3. *Developing fair recognition models.* The present focus for refining recognition models is still improving accuracy on specific datasets that are with very limited diversities considering gender, race, etc. For the future, many efforts should be paid to make the recognition models fairer. That is to reduce the differences in accuracy for different gender, ethnicity, etc. and bridge the gaps.

4. *Technical grounding of informed consent, data revocation and model retraining.* The understanding of privacy may change over time. Except for the maturation of recognition models, grounding of informed consent is essential to improve the acceptance of facial recognition for good (GDPR 2018; BAAI et al. 2019; NGPNGAI 2019). At the same time, since users may not really understand the items listed in informed consent notes, and may change their minds, data erasure and revocation mechanisms and models need to be realized and deployed (GDPR 2018; BAAI et al. 2019; NGPNGAI 2019). The real technological bottleneck is not on how data could be erased and revoked from the database, but about how to remove them from the training and prediction model, and how to retrain the recognition model with minimum cost.

5. *Developing privacy-preserving learning and recognition models.* One of the main focuses for facial and other biometric recognition is having access to privacy-related data for individuals. Future directions should be on the training and learning over encrypted data, and ensuring learning the general characteristics of the whole data without or at least reduce disclosing the private information of individuals.

6. *Ethics by Design and Risk Evaluation.* The research, development, and deployment of responsible facial and other biometric recognition to the society should be with ethics by design, and taking the considerations into the whole lifecycle of the services. In order to reduce the potential negative effects, risks evaluation mechanisms and platforms need to be developed and deployed. AI risks evaluation organizations need to be involved in the lifecycle to help find, point out, and reduce potential technical and ethical risks.